%% file: main.tex
\documentclass[10pt,twocolumn,letterpaper]{article}

\usepackage{cvpr}              

\input{preamble}

\makeatletter
\robustify\@latex@warning@no@line
\makeatother
\usepackage{authblk}
\definecolor{cvprblue}{rgb}{0.21,0.49,0.74}
\usepackage[pagebackref,breaklinks,colorlinks,citecolor=cvprblue]{hyperref}

\def\paperID{27} 
\def\confName{CVPR}
\def\confYear{2024}

\title{\OurMethod: Scaling Radiance Fields via Parameter Interpolation}

\makeatletter
\renewcommand\AB@affilsepx{ \quad \protect\Affilfont}
\makeatother
\begin{document}
\author[1]{Clinton Wang\thanks{}}
\author[2]{Peter Hedman}
\author[1]{Polina Golland}
\author[2]{Jonathan T. Barron}
\author[2]{Daniel Duckworth}
\affil[1]{MIT CSAIL}
\affil[2]{Google DeepMind}

\twocolumn[{
    \renewcommand\twocolumn[1][]{#1}
    \maketitle
    \vspace{-25pt}
    \captionsetup{type=figure}
    \input{figures/teaser}
    \vspace{15pt}
}]
{
  \renewcommand{\thefootnote}%
    {\fnsymbol{footnote}}
  \footnotetext[1]{Work done as a student researcher at Google.}
}
%
\begin{abstract}
Neural Radiance Fields (NeRFs) have unmatched fidelity on large, real-world scenes.
A common approach for scaling NeRFs is to partition the scene into regions, each of which is assigned its own parameters.
When implemented naively, such an approach is limited by poor test-time scaling and inconsistent appearance and geometry.
We instead propose \OurMethod{}, a novel architecture for rendering a target view using a subset of the model's parameters.
Our approach enables out-of-core training and rendering, increasing total model capacity with only a modest increase to training time.
We demonstrate significant improvements in multi-room scenes while remaining competitive on standard benchmarks.
\end{abstract}
\vspace{-4pt}

\section{Introduction}
\label{sec:intro}

Neural Radiance Fields (NeRFs) are a class of powerful, high-fidelity representations for 3D reconstruction achieving unparalleled reconstruction accuracy.
Unlike many approaches, NeRF models are straightforward to train and resilient to local minima while delivering impressive quality across a wide variety of scenes.
Nevertheless, reconstruction quality is inherently limited by model capacity, and new methods are needed to scale beyond the memory and compute limitations of present-day hardware.

A natural way to increase capacity is to spatially partition network parameters based on geometry~\cite{rebain2021derf,turki2022meganerf,wu2023scanerf,reiser2021kilonerf} or camera~\cite{tancik2022blocknerf} location.
Geometry-based partitioning divides the scene into multiple regions, which each have their own set of parameters.
While a fraction of model parameters are required to render a target ray, the number and choice of parameters varies depending on ray origin and direction and occlusions within the scene itself.
Camera-based partitioning, on the other hand, assigns parameters to possible query cameras.
While rendering is typically simpler and more efficient, inconsistencies arise when multiple parameter sets redundantly represent the same scene content.

We introduce \OurMethod, a high-capacity NeRF architecture tailored to large, multi-room scenes.
Key to our approach is the concept of \textit{camera-centric parameter interpolation}: the efficient interpolation of network parameters based on camera origin.
In particular, we designate a subset of model parameters as spatially-partitioned, loading and unloading parameters relevant to the active camera region during training.
At rendering time, the appropriate model parameters are loaded based on camera origin.
Given sufficient training time, our method outperforms a state-of-the-art baseline by a wide margin on the \ZipNeRF{} dataset.



\section{Related Work}
\label{sec:related}

\boldparagraph{Large-Scale 3D Reconstruction}
There is a large body of work tackling large scene reconstruction. This includes a number of classic works that apply structure from motion to extremely large photo collections \cite{agarwal2011building, schoenberger2016sfm}.
Ever since neural radiance fields \cite{mildenhall2020nerf} emerged as the dominant paradigm for novel view synthesis, a number of works have sought to also scale NeRF to large scenes. BungeeNeRF \cite{xiangli2022bungeenerf} combines satellite and ground level images in a single NeRF by appending residual blocks to represent progressively finer scales. Each block has its own output head that predicts residual color and density, but earlier blocks are only trained on coarser scale (aerial) images. Similar ideas for rendering at high resolution in a progressive manner are explored in variable bitrate neural fields and PyNeRF  \cite{takikawa2022variable, turki2023pynerf}.

Grid-Guided NeRFs \cite{xu2023gridguided} initially trains instant-NGP with a small MLP and then later incorporates a large view-dependent MLP, an approach that helps the model perform better on large urban scenes. F2-NeRF \cite{wang2023f2nerf} proposes adaptive space warping by finding local perspective warps that shrink regions far from any camera (a generalization of the NDC transform to multiple cameras), which helps to allocate model capacity more efficiently in scenes with long camera trajectories.
VR-NeRF \cite{VRNeRF} uses a customized camera rig to capture high resolution HDR footage that can be used to train a large-scale NeRF.

\boldparagraph{Submodel NeRFs}
Several works train independent NeRF submodels that are each responsible for representing a subset of the scene, and aggregate their predictions at inference time. Some works were motivated by efficiency or composition \cite{reiser2021kilonerf, rebain2020derf}. In \BlockNeRF \cite{tancik2022blocknerf}, street view data is represented by submodels centered at street intersections, each trained on cameras within a fixed radius. Target views are rendered by compositing whichever submodels have visibility of the region, as estimated by an auxiliary visibility network.
Mega-NeRF \cite{turki2022meganerf} partitions a scene using an octree, and trains each cell only on rays that pass through it.
Scalable Urban Dynamic Scenes \cite{turki2023suds} builds on this work by using LIDAR data to prune rays that do not terminate within the cell.
ScaNeRF \cite{wu2023scanerf} uses a similar tile-based submodel partition, and additionally performs bundle adjustment within a distributed parallel training framework.
Streamable MERF \cite{duckworth2024smerf} distills a large NeRF into several submodels that use a memory-efficient triplane representation. The work also features a deferred MLP whose parameters are interpolated based on the distance of the camera origin from the submodel centers, similar to the way parameters are interpolated in our work.

\section{Preliminaries}
\label{sec:preliminaries}

\input{figures/framework}

\boldparagraph{NeRF}
A \NeRF~\cite{mildenhall2020nerf} is a neural network that represents a 3D scene as a continuous mapping from a 3D spatial location $\PPosition_i$ to a volumetric density $\density_i$ and color $\Radiance_i$.
Novel images of the scene are rendered by tracing rays through this neural representation, where each ray $\ray(t) = \origin + t \, \viewdir$ is rendered according to the Gaussian quadrature approximation to the volume rendering equation,
\begin{gather}
    \rgb = \sum_i \Transmittance_i \lft( 1 - \exp\lft(-\density_i \delta_i \rgt) \rgt) \radiance_i \\
    \Transmittance_i = \exp \lft(\sum_{j=1}^{i-1} -\density_j \delta_j \rgt) 
        \text{ where } \delta_i = t_{i+1} - t_i \text{.}
    \label{eq:volume_rendering}
\end{gather}
Ray intervals are partitioned into non-overlapping intervals $\{ [t_i, t_{i+1}] \}$, and a multilayer perceptron (MLP) is trained to estimate $\density_i$ and $\radiance_i$ of these intervals from $\PPosition_i$ (taken to be the center of each interval) \cite{mildenhall2020nerf}.
This process is repeated, where a \emph{proposal network} learns to map a uniform sampling of distances $\{t_i\}$ to a ``coarse'' set of densities, which are then resampled to produce new distances that are concentrated around scene content~\cite{barron2022mipnerf360}.

\boldparagraph{\InstantNGP{} and \ZipNeRF{}}
\InstantNGP{}~\cite{muller2022instant} introduced a learned, multi-resolution datastructure for efficiently featurizing spatial coordinates throughout the scene. 
At coarse scales in the datastructure, spatial coordinate $\PPosition_i$ is trilinearly interpolated into a dense, multi-channel 3D grid to produce feature vectors (which we refer to as \emph{grid features}), and at fine scales $\PPosition_i$ is trilinearly interpolated into a 3D grid backed by a hash table to produce a feature vector (which we call \emph{hash features}).
\InstantNGP{} works by querying a spatial coordinate across all scales and concatenating the interpolated outputs to construct a feature vector.
We write this interpolation and concatenation as: $\feature_i = \ingp (\mathbf{x}_i)$.
A small MLP called the ``geometry MLP'' then takes these feature vectors and predicts a scalar density value $\density_i$, and a larger MLP takes both $\feature_i$ and the viewing direction of the ray $\viewdir$ as input to predict radiance $\radiance_i = \mlp(\feature_i, \viewdir)$.

Building on mip-NeRF \cite{barron2021mipnerf} and \InstantNGP{}, \ZipNeRF \cite{barron2023zipnerf} uses multisampling and reweighting to parameterize the conical sub-frusta along each ray using NGPs. 
This strategy achieves the anti-aliasing and scale-reasoning provided by mip-NeRF while being significantly faster to optimize and evaluate. \ZipNeRF still attains the highest quality in novel view synthesis, outperforming 3D Gaussian Splatting \cite{gaussiansplats3d} on photos acquired with rectilinear lenses and additionally able to accommodate distorted lenses (which 3DGS cannot).

\section{\OurMethod{}}
\label{sec:method}

We propose Interpolated NeRF (\OurMethod{}), a scalable, out-of-core approach for increasing model capacity without a corresponding increase in memory usage.
We adopt the concept of \emph{spatially-partitioned} model parameters: each parameter is broadly categorized as either \PARTITIONED{} or \SHARED{}, with \PARTITIONED{} parameters varying with camera origin and \SHARED{} remaining unchanged.
To render a particular camera, we interpolate \PARTITIONED{} parameters within a local neighborhood and perform a forward pass similar to typical NeRF model.

While interpolated parameters provide the model the freedom to specialize to relevant camera views, they also permit origin-dependent geometry, and geometric coherence can suffer when all parameters are interpolated.
Thus the role of the \SHARED{} parameters is to maintain a stable coarse geometry, and we choose to share proposal network parameters and all NGP grid features.

\boldparagraph{Parameter grid}
To determine the anchor locations for parameter sets, we begin by establishing a 2D axis-aligned bounding box containing all training cameras.
The bounding box is partitioned into $\NumSubmodelsX \times \NumSubmodelsY$ non-isotropic grid cells.
We assume that scenes are largely planar and thus omit a partitioning along the vertical axis.
Cameras are then assigned to cells based on their origin (see \cref{fig:framework}).
Note that only cells with a sufficient number of training cameras are instantiated; if a camera's origin lies outside of an instantiated cell, cameras are reassigned to the nearest active cell instead.

\boldparagraph{Parameter mixing}
Similar to \BlockNeRF{}, we assign each vertex in the parameter grid its own \emph{parameter set} for interpolation.
However, \BlockNeRF{} applies nearest neighbor interpolation, which introduces discontinuities at the boundaries between parameter sets and necessitates post-processing heuristics like visibility maps and image-space interpolation to avoid ``popping'' artifacts.
We instead explore \emph{bilinear} interpolation, which naturally results in a continuous field of model parameters.

We implement bilinear interpolation layer-by-layer, performing a forward pass of each layer with four parameter sets followed by interpolation of their outputs.
In the NGP component, four sets of hash features are retrieved and interpolated at each resolution; in the MLP, each fully connected layer performs four multiply-adds followed by interpolation.
This approach avoids constructing per-sample model parameters, which would significantly increase memory usage. 

\boldparagraph{Cell-by-cell training}
To decouple memory usage from parameter count, we optimize four parameter sets at a time corresponding to a single grid cell.
We optimize cells serially in a round robin fashion, loading and unloading partitioned model parameters as needed.
When optimizing a cell, we by default use training cameras that lie within the same cell.
Rather than optimizing cells for a fixed number of iterations, we vary the number of iterations proportionally to the number of cameras assigned to it.
We find this improves quality and reduces floating artifacts.

\boldparagraph{Ray reassignment}
To increase the amount of training signal each cell receives, we incorporate cameras from other cells during optimization.
In particular, each training batch is constructed such that a fixed percentage $\ProbRayReassign$ of camera rays are sourced from neighbors at most $\KRayReassign$ cells away. The interpolation weights are determined by projecting the camera origin to the closest point in the cell.
We find that this strategy strongly reduces floating geometry immediately outside of the training camera frustum.

\section{Experiments}
\label{sec:experiments}

\input{tables/zipnerf_ablations}
\input{tables/mipnerf360_ablations}
\input{tables/mipnerf360_headline}
\input{figures/comparison}

\boldparagraph{Datasets}
We evaluate our approach on two datasets: four large multi-room scenes introduced by \ZipNeRF{}~\cite{barron2023zipnerf} and the four indoor and five outdoor scenes of the \MipNeRF dataset~\cite{barron2022mipnerf360}.
All models were trained from scratch using the original, distorted photos with every 8th image held out for test.
To illustrate the higher fidelity of our approach, we use images at double their typical resolution: $1752 \times 1168$ for \ZipNeRF{} and $2456 \times 1632$ or $3120 \times 2080$ for \MipNeRF{}.
All camera parameters have been estimated with COLMAP at full resolution~\cite{schoenberger2016sfm}.
We note that this dataset is not amenable rasterization methods such as 3D Gaussian Splatting, which assume a pinhole camera model.

\boldparagraph{Baseline}
We compare our method to \ZipNeRF{}, a state-of-the-art NeRF method for reconstruction of large indoor spaces.
Our implementation uses three networks: two for proposing ray intervals and one for color and density prediction.
All three networks consist of a multiresolution hash grid with up to $2^{21}$ entries followed by a per-interval geometry MLP with 1 layer and 64 units.
The last network additionally employs an appearance MLP with 4 layers and 256 units for predicting view-dependent color.

\boldparagraph{Training Details}
Outside of the parameter partitioning scheme proposed in this work, our model architecture matches that of the \ZipNeRF{} baseline.
In experiments on the \ZipNeRF{} dataset, we partition all parameters in the second proposal network and final density and appearance network and set $\ProbRayReassign=0.3$, $\KRayReassign=1$.
In experiments on the \MipNeRF{} dataset, only the final network's parameters are partitioned and we set $\ProbRayReassign=0.8$, $\KRayReassign=4$.

To maintain a similar memory footprint with the baseline, we reduce the number of multiresolution hash grid entries per partitioned variable by $4\times$.
This ensures that our method uses roughly as much device memory as the baseline, even if the \emph{total} number of parameters is larger.

\boldparagraph{Comparison to Baseline}
On the \ZipnerfDataset, our method achieves significantly higher quality than the \ZipNeRF{} baseline at 400,000 steps; see \cref{tab:zipnerf-ablations}.
Qualitatively, this translates to higher geometric and texture detail; see \cref{fig:comparison_zipnerf}.
We further observe a clear, positive relationship between model quality, training time, and number of cells.
In contrast, baseline quality does \emph{not} improve with training time.
This strongly suggests model capacity, rather than optimization time, to be \ZipNeRF{}'s limiting factor.

We perform a similar study on a subset of the \MipNeRFThreeSixty{} dataset, where a similar relationship is lacking; see Tables~\ref{tab:mipnerf_ablation} and \ref{tab:mipnerf360-headline}.
While our method modestly outperforms the baseline at 200,000 steps in terms of PSNR and SSIM, a negative relationship between LPIPS and number of cells is evident, as each parameter set receives fewer training iterations as the number of parameter sets grows under a fixed total training budget.
This study further suggests that \ZipNeRF{}'s model capacity is sufficient for medium-sized scenes.





\boldparagraph{Runtime Analysis}
Although our method achieves higher reconstruction quality, the current implementation trains more slowly than its \ZipNeRF{} counterpart.
The reason for this is twofold: first, our method applies each partitioned layer four times to the baseline's one; second, partitioned parameters are swapped in and out of device memory during training.
Based on back-of-the-envelope calculations, we believe the time required for the latter to be largely avoidable with appropriate optimizations.
We further believe the number of training steps can be significantly reduced by taking advantage of model parallelism, assigning each cell to a different device.
We leave the design and implementation of both of these directions to future work.

\section{Conclusion}
\label{sec:conclusion}

In this work, we have introduced \OurMethod{}, a scalable, out-of-core NeRF model architecture for reconstructing large, multi-room scenes.
We demonstrated that parameter interpolation is an effective approach for increasing model capacity without a corresponding increase to memory or compute requirements.
While our method demonstrates impressive quality, additional work is needed to reduce training time, compare our approach to other submodel-based approaches, test other interpolation schemes, and validate our method on even larger (e.g. city-scale) scenes. 

\paragraph{Acknowledgments} We would like to thank Suhani Vora for providing logistical support.

{
    \small
    \bibliographystyle{ieeenat_fullname}
    \bibliography{main}
}


\clearpage
\maketitlesupplementary

\section{Training Details}
\label{sec:training-details}

\boldparagraph{Training}
We implement our method and the \ZipNeRF{} baseline by building on the \href{https://github.com/jonbarron/camp_zipnerf}{\texttt{camp\_zipnerf}} codebase.
Models are trained using sixteen A100s for up to 24 hours.
We optimize all models with a batch size of $2^{16}$ rays using the Adam optimizer.
Unless otherwise stated, hyperparameters match those described in the \ZipNeRF{} paper~\cite{barron2023zipnerf}.
We employ an exponentially decaying learning rate schedule with an initial learning rate of 1e-2 and a final learning rate of 1e-3 or 1e-4 on the \ZipNeRF{} and \MipNeRF{} datasets, respectively.
We further introduce a linear learning rate warm-up schedule of 2,500 steps starting with an initial learning rate of 1e-8.
We apply $L_2$ regularization to all hash grid parameters with a weight of 0.001 for proposal networks and 0.1 for density and appearance networks on the \ZipNeRF{} dataset and 0.1 for all networks on the \MipNeRF{} dataset.
We further force volumetric rendering weights along each camera ray to sum to unity.

\boldparagraph{Model Architecture}
All models, including the \ZipNeRF{} baseline and \OurMethod{}, are composed of three networks: two proposal networks and one appearance and density network.
Each network, in turn, is composed of up to three components: a multiresolution hash grid, a Geometry MLP, and potentially an Appearance MLP.
All three networks contain the first two components; only the last contains an Appearance MLP.
The shape of these MLPs is identical for all methods as described in the main text.

The majority of each method's parameters lie in their multiresolution hash grid.
We describe the number of hash resolution levels, number of entries per level, number of features per entry, and spatial resolution of these multiresolution grids in Tables~\ref{tab:model-size-mipnerf360} and \ref{tab:model-size-zipnerf}.
We further indicate which multiresolution grids are spatially-partitioned and interpolated for our method.

\begin{table}
    \newcommand{\verticaltext}[2]{
      \parbox[t]{1mm}{\multirow{#1}{*}{\rotatebox[origin=c]{90}{#2}}} 
    }
    \centering
    \resizebox{\linewidth}{!}{
    \begin{tabular}{l|c|c|c|c|c|c}
        & Network & Levels & Buckets & Features & Interp. & Resol. \\
        \hline
        \verticaltext{3}{Base.}
        & \texttt{Prop1} & 6 & $2^{21}$ & 1 & No & 512 \\ 
        & \texttt{Prop2} & 8 & $2^{21}$ & 1 & No & 2048 \\ 
        & \texttt{Final} & 10 & $2^{21}$ & 4 & No & 8192 \\ 
        \hline
        \verticaltext{3}{Ours}
        & \texttt{Prop1} & 6 & $2^{21}$ & 1 & No & 512 \\ 
        & \texttt{Prop2} & 8 & $2^{21}$ & 1 & No & 2048 \\ 
        & \texttt{Final} & 10 & $2^{19}$ & 4 & Yes & 8192 \\ 
    \end{tabular}   
    }
    \caption{Multiresolution hash grid sizes on the \MipNeRF{} dataset.}
    \label{tab:model-size-mipnerf360}
\end{table}

\begin{table}
    \newcommand{\verticaltext}[2]{
      \parbox[t]{1mm}{\multirow{#1}{*}{\rotatebox[origin=c]{90}{#2}}} 
    }
    \centering
    \resizebox{\linewidth}{!}{
    \begin{tabular}{l|c|c|c|c|c|c|c}
        & Network & Levels & Buckets & Features & Interp. & Resol. \\
        \hline
        \verticaltext{3}{Base.}
        & \texttt{Prop1} & 6 & $2^{21}$ & 1 & No & 512 \\ 
        & \texttt{Prop2} & 8 & $2^{21}$ & 1 & No & 2048 \\ 
        & \texttt{Final} & 10 & $2^{21}$ & 4 & No & 8192 \\ 
        \hline
        \verticaltext{3}{Ours}
        & \texttt{Prop1} & 7 & $2^{21}$ & 1 & No & 1024 \\ 
        & \texttt{Prop2} & 9 & $2^{19}$ & 1 & Yes & 4096 \\ 
        & \texttt{Final} & 11 & $2^{19}$ & 4 & Yes & 16384 \\ 
    \end{tabular}   
    }
    \caption{Multiresolution hash grid sizes on the \ZipNeRF{} dataset.}
    \label{tab:model-size-zipnerf}
\end{table}

\boldparagraph{Cell Swapping}
We train our method by swapping \PARTITIONED{} model parameters in and out of device memory as needed.
As described in \cref{sec:method}, scene grid cells are optimized one at a time.
We optimize each cell $k$ for $2 \times N_k$ iterations, where $N_k$ is the number of training cameras allocated to this cell. 
When a new grid cell is chosen, \SHARED{} parameters are left in memory while the appropriate \PARTITIONED{} are swapped in.
To order cells, we assign each cell a linearized integer identifier based on its position in the 2D scene grid and loop over cells according to this identifier.
When loading a new cell for training, \PARTITIONED{} variables along with associated Adam statistics are loaded into memory.

\boldparagraph{Dataset}
Unlike the majority of prior work, we use high-resolution versions of the \MipNeRF{} and \ZipNeRF{} datasets with the original lens distortion.
This enables a deeper exploration of model capacity -- a key goal of this work -- but prevents metrics from being directly comparable to prior publications. 
The use of distorted photos further limits our ability to compare to rasterization-based methods such as 3D Gaussian Splatting~\cite{gaussiansplats3d}.
Concretely, we use photos at double the resolution used in prior work.
For the \MipNeRF{} dataset, this means full-resolution photos for indoor scenes and $2 \times$ downsampled photos for outdoor scenes.
For the \ZipNeRF{} dataset, we use $2 \times$ downsampled photos for all scenes except for \textsc{Berlin}, where we use $4 \times$ downsampling.


\section{Additional Results}

\input{figures/gardenvase_compare}
\input{figures/qualitative}

Overall, we find a strong positive correlation between model capacity and texture detail on multi-room scenes as demonstrated in \cref{fig:qualitative-zipnerf}.
Our method consistently provides crisp detail on textured surfaces including painting, carpets, and curtains.
The same cannot reliably be said, however, for the \MipNeRF{} dataset.
In the majority of scenes, our method is indistinguishable from the \ZipNeRF{} baseline as in \cref{fig:gardenvase-compare}.

\end{document}

%% file: preamble.tex
\usepackage[table,x11names,dvipsnames]{xcolor}
\usepackage{multirow}

\definecolor{turquoise}{cmyk}{0.65,0,0.1,0.3}
\definecolor{purple}{rgb}{0.65,0,0.65}
\definecolor{dark_green}{rgb}{0, 0.5, 0}
\definecolor{orange}{rgb}{0.8, 0.6, 0.2}
\definecolor{red}{rgb}{0.8, 0.2, 0.2}
\definecolor{darkgray}{rgb}{0.5, 0.5, 0.5}
\definecolor{darkred}{rgb}{0.6, 0.1, 0.05}
\definecolor{blueish}{rgb}{0.0, 0.3, .6}
\definecolor{light_gray}{rgb}{0.7, 0.7, .7}
\definecolor{pink}{rgb}{1, 0, 1}
\definecolor{greyblue}{rgb}{0.15, 0.25, 0.65}

\newcommand{\copydanger}[1]{\textbf{OMITTED COPY/PASTE TEXT}}
\newcommand{\ignore}[1]{}

\renewcommand{\paragraph}[1]{\vspace{.5em}\noindent\textbf{#1}.}

\newcommand{\PSNR}{PSNR\xspace}
\newcommand{\SSIM}{SSIM\xspace}
\newcommand{\LPIPS}{LPIPS\xspace}

\newcommand{\ZipNeRF}{Zip-NeRF\xspace}
\newcommand{\MipNeRFThreeSixty}{mip-NeRF 360\xspace}
\newcommand{\OurMethod}{InterNeRF\xspace}

\newcommand{\NeRF}{NeRF\xspace}
\newcommand{\Ours}{Ours}
\newcommand{\MipNeRF}{mip-NeRF 360\xspace}
\newcommand{\BlockNeRF}{Block-NeRF\xspace}

\newcommand{\InstantNGP}{Instant-NGP\xspace}
\newcommand{\SupplementaryMaterial}[1]{{supplementary material}}
\newcommand{\ZipnerfDataset}{Zip-NeRF dataset\xspace}

\usepackage{enumitem}
\setlist[itemize]{noitemsep,leftmargin=*,topsep=0em}
\setlist[enumerate]{noitemsep,leftmargin=*,topsep=0em}

\newcommand{\density}{\tau}

\newcommand{\feature}{\mathbf{z}}

\newcommand{\lft}{\mathopen{}\left}

\newcommand{\origin}{\mathbf{o}}

\newcommand{\pposition}{\mathbf{x}}
\newcommand{\radiance}{\mathbf{c}}
\newcommand{\ray}{\mathbf{r}}

\newcommand{\rgb}{\mathbf{C}}
\newcommand{\rgt}{\aftergroup\mathclose\aftergroup{\aftergroup}\right}

\newcommand{\viewdir}{\mathbf{d}}

\newcommand{\PPosition}{\pposition}

\newcommand{\Radiance}{\radiance}

\newcommand{\Transmittance}{\mathrm{T}}

\newcommand{\NumSubmodelsX}{N_x}
\newcommand{\NumSubmodelsY}{N_y}

\newcommand{\PARTITIONED}{\textsc{interpolated}\xspace}
\newcommand{\SHARED}{\textsc{shared}\xspace}
\newcommand{\ProbRayReassign}{p_{\text{r}}}
\newcommand{\KRayReassign}{K_{\text{r}}}

\newcommand{\ingp}{\mathrm{NGP}\xspace}
\newcommand{\mlp}{\mathrm{MLP}\xspace}

\newcommand{\boldparagraph}[1]{\vspace{0.1cm}\noindent{\textbf{#1:}}\phantomsection}

\newcommand{\appendixref}[1]{the appendices}
\newcommand{\appendixtableref}[1]{the appendices}

\newcommand{\CaptionToTableVspace}{\vspace{-0.05in}}
\newcommand{\AfterTableVspace}{\vspace{0.01in}}

%% file: figures/teaser.tex
\centering
\includegraphics[width=.95\textwidth]{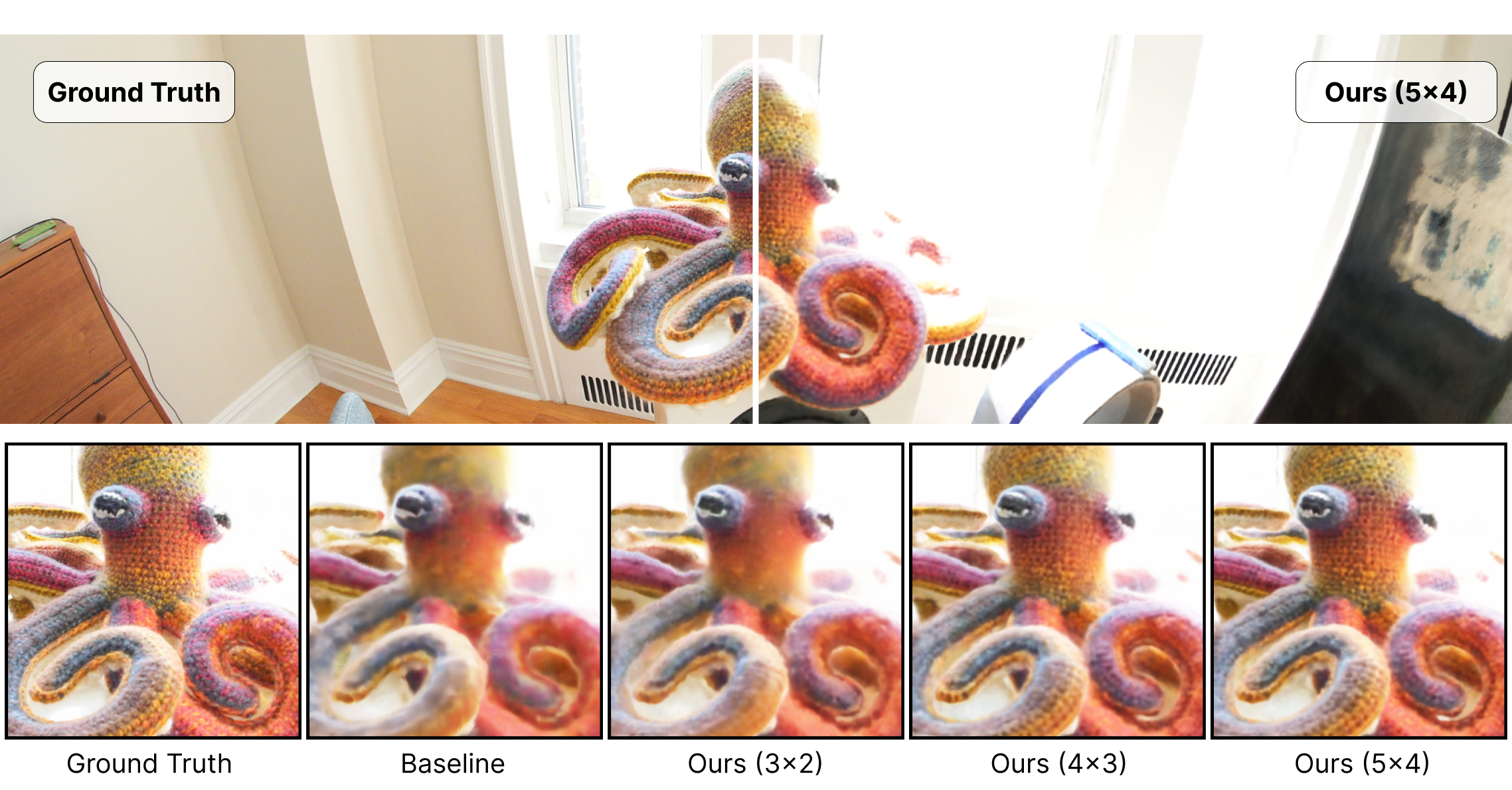}
\vspace{-4pt}
\caption{
    \OurMethod{} achieves state-of-the-art reconstruction quality on large, multi-room scenes via parameter interpolation.
    Parameters are anchored to locations in space and interpolated based on camera position.
    As a result, \OurMethod{} achieves high geometric and texture accuracy at centimeter scale.
}
\label{fig:teaser}

%% file: figures/framework.tex
\begin{figure*}[htb]
  \centering
  \includegraphics[width=.94\textwidth]{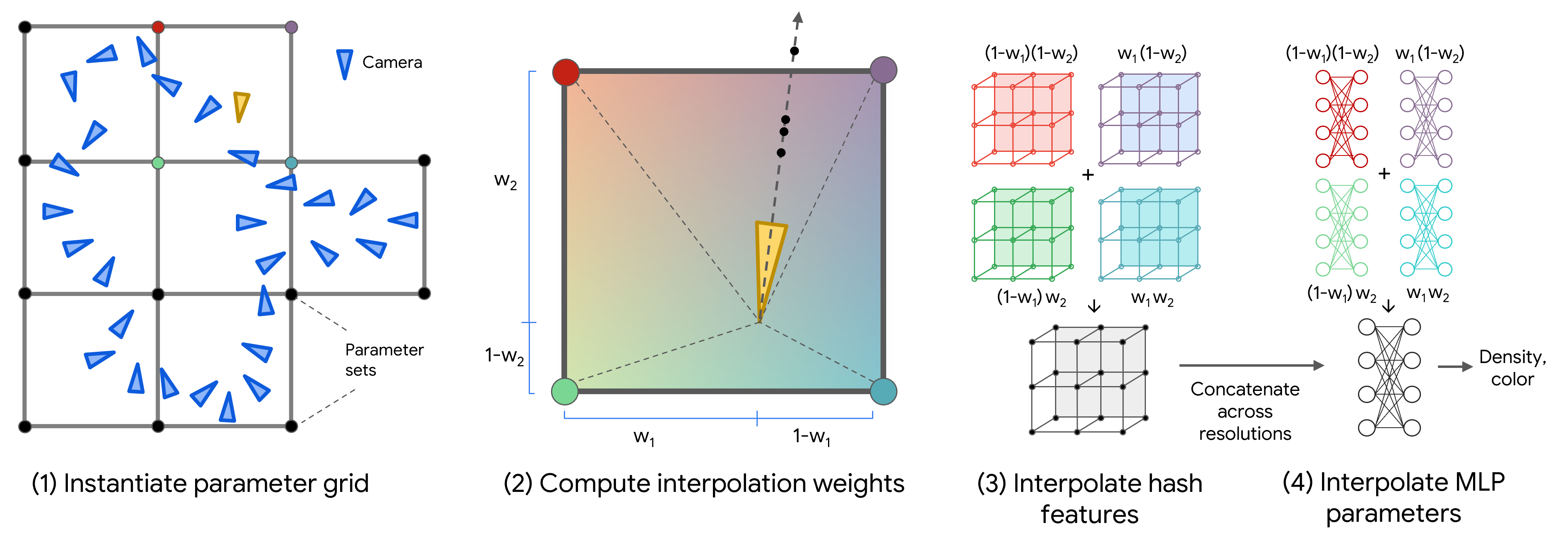}
\vspace{-7pt}
  \caption{
  \textbf{Our framework}.
  (1) We partition the scene into a parameter grid and assign training cameras to each cell based on its origin.
  (2) For a given (training or target) view, we obtain mixing weights as the bilinear interpolation coefficients based on camera origin.
  (3) Each query point along the ray is used to index into a multi-resolution set of grid features per parameter set, with either explicit assignment or a hash table.
  The mixing weights are applied here to yield a single set of features.
  (4) Each parameter set also has its own set of MLP weights, which are combined using the same mixing weights to form a new MLP.
  }
  \label{fig:framework}
\end{figure*}

%% file: tables/zipnerf_ablations.tex
\begin{table*}[ht!]
\centering
\resizebox{\textwidth}{!}{
\begin{tabular}{c|cccc|cccc|cccc|c}
 & \multicolumn{4}{c}{\PSNR{}} & \multicolumn{4}{|c}{\SSIM{}} & \multicolumn{4}{|c}{\LPIPS{}} & \multicolumn{1}{|c}{Wall Time}\\
Method / Steps & 50K & 100K & 200K & 400K & 50K & 100K & 200K & 400K & 50K & 100K & 200K & 400K & @ 400K \\
\hline
\ZipNeRF{} & {\cellcolor[rgb]{1.000, 1.000, 1.000} 25.71} & {\cellcolor[rgb]{1.000, 1.000, 1.000} 25.82} & {\cellcolor[rgb]{1.000, 1.000, 1.000} 25.84} & {\cellcolor[rgb]{1.000, 1.000, 1.000} 25.83} & {\cellcolor[rgb]{1.000, 1.000, 1.000} 0.796} & {\cellcolor[rgb]{1.000, 1.000, 1.000} 0.799} & {\cellcolor[rgb]{1.000, 1.000, 1.000} 0.800} & {\cellcolor[rgb]{1.000, 1.000, 1.000} 0.799} & {\cellcolor[rgb]{1.000, 1.000, 1.000} 0.373} & {\cellcolor[rgb]{1.000, 1.000, 1.000} 0.368} & {\cellcolor[rgb]{1.000, 1.000, 1.000} 0.365} & {\cellcolor[rgb]{1.000, 1.000, 1.000} 0.366} & 6.8h \\
\Ours{} (3$\times$2) & {\cellcolor[rgb]{1.000, 1.000, 1.000} 25.83} & {\cellcolor[rgb]{1.000, 1.000, 1.000} 25.98} & {\cellcolor[rgb]{1.000, 1.000, 1.000} 26.07} & {\cellcolor[rgb]{1.000, 0.958, 0.937} 26.14} & {\cellcolor[rgb]{1.000, 1.000, 1.000} 0.801} & {\cellcolor[rgb]{1.000, 1.000, 1.000} 0.803} & {\cellcolor[rgb]{1.000, 1.000, 1.000} 0.805} & {\cellcolor[rgb]{1.000, 1.000, 1.000} 0.805} & {\cellcolor[rgb]{1.000, 1.000, 1.000} 0.372} & {\cellcolor[rgb]{1.000, 1.000, 1.000} 0.367} & {\cellcolor[rgb]{1.000, 1.000, 1.000} 0.363} & {\cellcolor[rgb]{1.000, 1.000, 1.000} 0.362} & 17.5h \\
\Ours{} (4$\times$3) & {\cellcolor[rgb]{1.000, 1.000, 1.000} 25.95} & {\cellcolor[rgb]{0.998, 0.922, 0.886} 26.24} & {\cellcolor[rgb]{0.996, 0.878, 0.823} 26.35} & {\cellcolor[rgb]{0.997, 0.894, 0.845} 26.31} & {\cellcolor[rgb]{1.000, 1.000, 1.000} 0.809} & {\cellcolor[rgb]{0.998, 0.912, 0.871} 0.815} & {\cellcolor[rgb]{0.996, 0.873, 0.817} 0.818} & {\cellcolor[rgb]{0.995, 0.864, 0.805} 0.818} & {\cellcolor[rgb]{1.000, 1.000, 1.000} 0.358} & {\cellcolor[rgb]{0.999, 0.930, 0.897} 0.346} & {\cellcolor[rgb]{0.997, 0.894, 0.845} 0.341} & {\cellcolor[rgb]{0.995, 0.864, 0.805} 0.339} & 19.9h \\
\Ours{} (5$\times$4) & {\cellcolor[rgb]{1.000, 1.000, 1.000} 25.61} & {\cellcolor[rgb]{0.997, 0.904, 0.860} 26.28} & {\cellcolor[rgb]{0.994, 0.832, 0.762} 26.42} & {\cellcolor[rgb]{0.988, 0.672, 0.561} 26.65} & {\cellcolor[rgb]{1.000, 1.000, 1.000} 0.806} & {\cellcolor[rgb]{0.995, 0.860, 0.799} 0.818} & {\cellcolor[rgb]{0.988, 0.732, 0.630} 0.823} & {\cellcolor[rgb]{0.988, 0.672, 0.561} 0.826} & {\cellcolor[rgb]{1.000, 1.000, 1.000} 0.358} & {\cellcolor[rgb]{0.997, 0.891, 0.842} 0.341} & {\cellcolor[rgb]{0.989, 0.746, 0.648} 0.331} & {\cellcolor[rgb]{0.988, 0.672, 0.561} 0.326} & 23.4h \\

\hline
\end{tabular}
}
\CaptionToTableVspace{}
\caption{
    Quantitative evaluation on the \ZipnerfDataset.
    Quality increases monotonically with scene grid resolution and training steps for our method while \ZipNeRF{} saturates.
    To maintain comparable memory usage, we provide \ZipNeRF{} with $4 \times$ more hash grid parameters than our interpolated models.
    \label{tab:zipnerf-ablations}
}
\AfterTableVspace{}
\end{table*}

%% file: tables/mipnerf360_ablations.tex
\begin{table*}[ht!]
\centering
\resizebox{\textwidth}{!}{
\begin{tabular}{c|cccc|cccc|cccc|c}
 & \multicolumn{4}{c}{\PSNR{}} & \multicolumn{4}{|c}{\SSIM{}} & \multicolumn{4}{|c}{\LPIPS{}} & \multicolumn{1}{|c}{Wall Time} \\
Method / Steps & 25K & 50K & 100K & 200K & 25K & 50K & 100K & 200K & 25K & 50K & 100K & 200K & @ 200K \\
\hline
\ZipNeRF{} & {\cellcolor[rgb]{1.000, 1.000, 1.000} 28.15} & {\cellcolor[rgb]{0.997, 0.894, 0.845} 28.28} & {\cellcolor[rgb]{0.994, 0.837, 0.769} 28.34} & {\cellcolor[rgb]{0.994, 0.846, 0.781} 28.34} & {\cellcolor[rgb]{0.994, 0.846, 0.781} 0.808} & {\cellcolor[rgb]{0.989, 0.746, 0.648} 0.813} & {\cellcolor[rgb]{0.988, 0.682, 0.572} 0.816} & {\cellcolor[rgb]{0.988, 0.677, 0.566} 0.816} & {\cellcolor[rgb]{0.995, 0.851, 0.787} 0.313} & {\cellcolor[rgb]{0.990, 0.764, 0.672} 0.305} & {\cellcolor[rgb]{0.988, 0.702, 0.595} 0.299} & {\cellcolor[rgb]{0.988, 0.672, 0.561} 0.296} & 2.7h \\
\Ours{} (2$\times$2) & {\cellcolor[rgb]{0.998, 0.927, 0.893} 28.23} & {\cellcolor[rgb]{0.993, 0.828, 0.756} 28.35} & {\cellcolor[rgb]{0.989, 0.755, 0.660} 28.42} & {\cellcolor[rgb]{0.989, 0.746, 0.648} 28.42} & {\cellcolor[rgb]{0.999, 0.935, 0.904} 0.801} & {\cellcolor[rgb]{0.994, 0.832, 0.762} 0.808} & {\cellcolor[rgb]{0.990, 0.760, 0.666} 0.812} & {\cellcolor[rgb]{0.988, 0.712, 0.607} 0.815} & {\cellcolor[rgb]{0.999, 0.948, 0.923} 0.327} & {\cellcolor[rgb]{0.996, 0.873, 0.817} 0.315} & {\cellcolor[rgb]{0.992, 0.796, 0.714} 0.308} & {\cellcolor[rgb]{0.989, 0.755, 0.660} 0.304} & 5.2h \\
\Ours{} (3$\times$3) & {\cellcolor[rgb]{1.000, 1.000, 1.000} 28.08} & {\cellcolor[rgb]{0.995, 0.851, 0.787} 28.33} & {\cellcolor[rgb]{0.989, 0.746, 0.648} 28.43} & {\cellcolor[rgb]{0.988, 0.672, 0.561} 28.49} & {\cellcolor[rgb]{1.000, 1.000, 1.000} 0.795} & {\cellcolor[rgb]{0.995, 0.855, 0.793} 0.807} & {\cellcolor[rgb]{0.988, 0.727, 0.624} 0.814} & {\cellcolor[rgb]{0.988, 0.672, 0.561} 0.817} & {\cellcolor[rgb]{1.000, 1.000, 1.000} 0.340} & {\cellcolor[rgb]{0.998, 0.917, 0.878} 0.322} & {\cellcolor[rgb]{0.994, 0.837, 0.769} 0.312} & {\cellcolor[rgb]{0.990, 0.773, 0.684} 0.306} & 7.1h \\
\Ours{} (4$\times$4) & {\cellcolor[rgb]{1.000, 1.000, 1.000} 27.86} & {\cellcolor[rgb]{0.999, 0.943, 0.915} 28.21} & {\cellcolor[rgb]{0.992, 0.805, 0.726} 28.38} & {\cellcolor[rgb]{0.988, 0.712, 0.607} 28.46} & {\cellcolor[rgb]{1.000, 1.000, 1.000} 0.779} & {\cellcolor[rgb]{1.000, 1.000, 1.000} 0.797} & {\cellcolor[rgb]{0.993, 0.828, 0.756} 0.809} & {\cellcolor[rgb]{0.988, 0.717, 0.613} 0.814} & {\cellcolor[rgb]{1.000, 1.000, 1.000} 0.363} & {\cellcolor[rgb]{1.000, 1.000, 1.000} 0.338} & {\cellcolor[rgb]{0.998, 0.914, 0.875} 0.322} & {\cellcolor[rgb]{0.994, 0.837, 0.769} 0.312} & 10.7h \\
\hline
\end{tabular}
}
\CaptionToTableVspace{}
\caption{
    Quantitative evaluation on four scenes from the \MipNeRFThreeSixty{} dataset: \textsc{bicycle}, \textsc{garden}, \textsc{counter}, and \textsc{bonsai}.
    PSNR and SSIM are competitive with \ZipNeRF{} as the number of training steps increases.
    Here we again provide \ZipNeRF{} with $4 \times$ more hash grid parameters than our interpolated models.
    \label{tab:mipnerf_ablation}
}
\AfterTableVspace{}
\end{table*}

%% file: tables/mipnerf360_headline.tex
\begin{table}
\centering
\small

\begin{tabular}{c|rrr|r}
                        & \PSNR{}                                      & \SSIM{}                                      & \LPIPS{}                                     & Wall Time \\
\hline
\ZipNeRF{}              & {\cellcolor[rgb]{1.000, 1.000, 1.000} 27.55} & {\cellcolor[rgb]{1.000, 1.000, 1.000} 0.794} & {\cellcolor[rgb]{0.988, 0.672, 0.561} 0.292} & {\cellcolor[rgb]{0.988, 0.672, 0.561} 2.73 Hr} \\
\Ours{} (2x2)           & {\cellcolor[rgb]{1.000, 1.000, 1.000} 27.66} & {\cellcolor[rgb]{1.000, 1.000, 1.000} 0.795} & {\cellcolor[rgb]{1.000, 1.000, 1.000} 0.299} & 5.21 Hr \\
\Ours{} (3x3)           & {\cellcolor[rgb]{0.988, 0.672, 0.561} 27.69} & {\cellcolor[rgb]{0.988, 0.672, 0.561} 0.796} & {\cellcolor[rgb]{1.000, 1.000, 1.000} 0.300} & 7.12 Hr \\
\Ours{} (4x4)           & {\cellcolor[rgb]{1.000, 1.000, 1.000} 27.64} & {\cellcolor[rgb]{1.000, 1.000, 1.000} 0.793} & {\cellcolor[rgb]{1.000, 1.000, 1.000} 0.307} & 10.68 Hr \\
\hline
\end{tabular}

\caption{
    Quantitative evaluation on the full \MipNeRFThreeSixty{} dataset.
    Models are trained for 200K steps.
}
\label{tab:mipnerf360-headline}
\end{table}

%% file: figures/comparison.tex
\begin{figure}[tb]
  \centering
  \begin{subfigure}{\linewidth}
  \includegraphics[width=\textwidth]{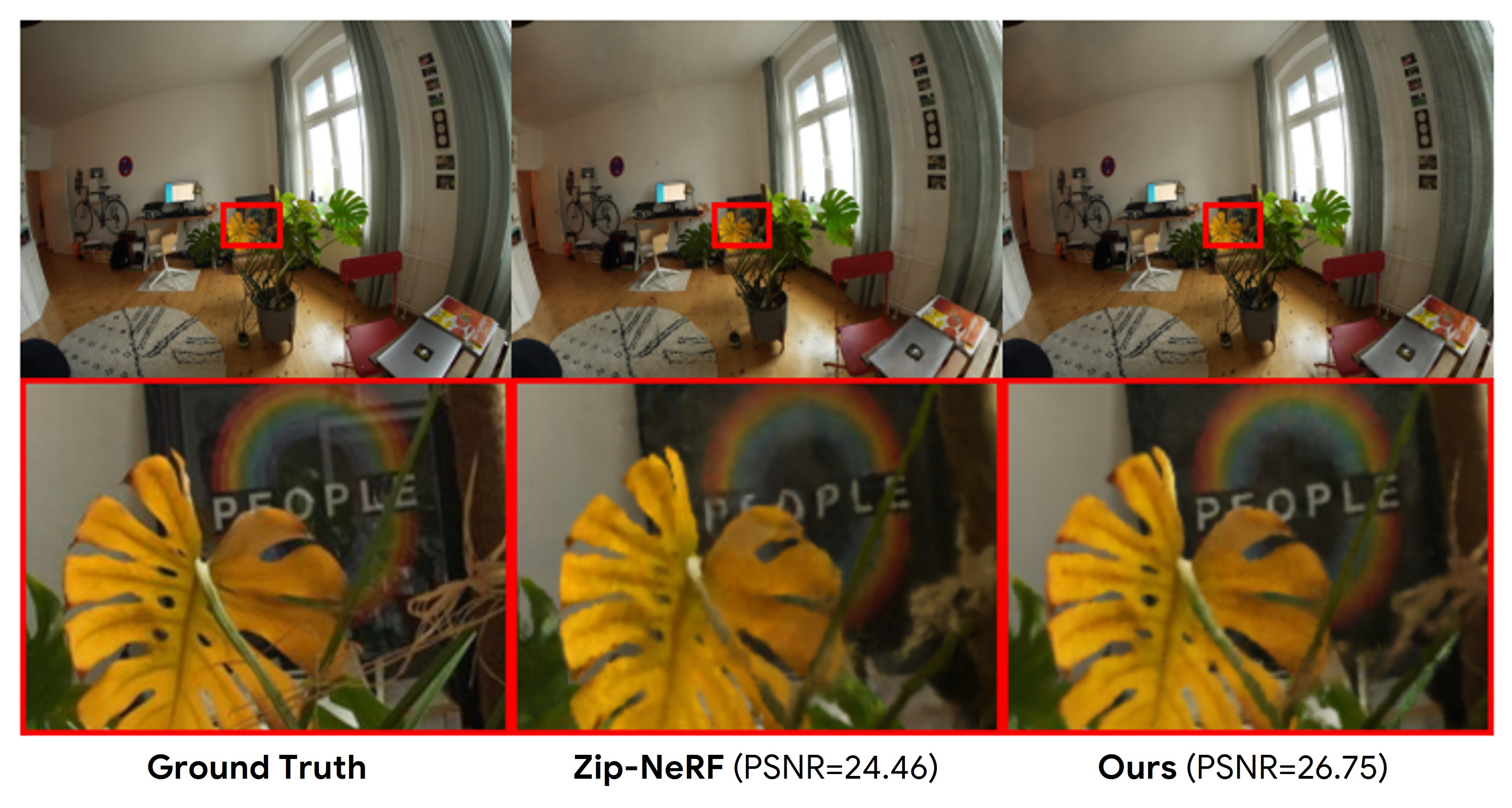}
  \end{subfigure}
  \vspace{-4pt}
  \begin{subfigure}{\linewidth}
  \includegraphics[width=\textwidth]{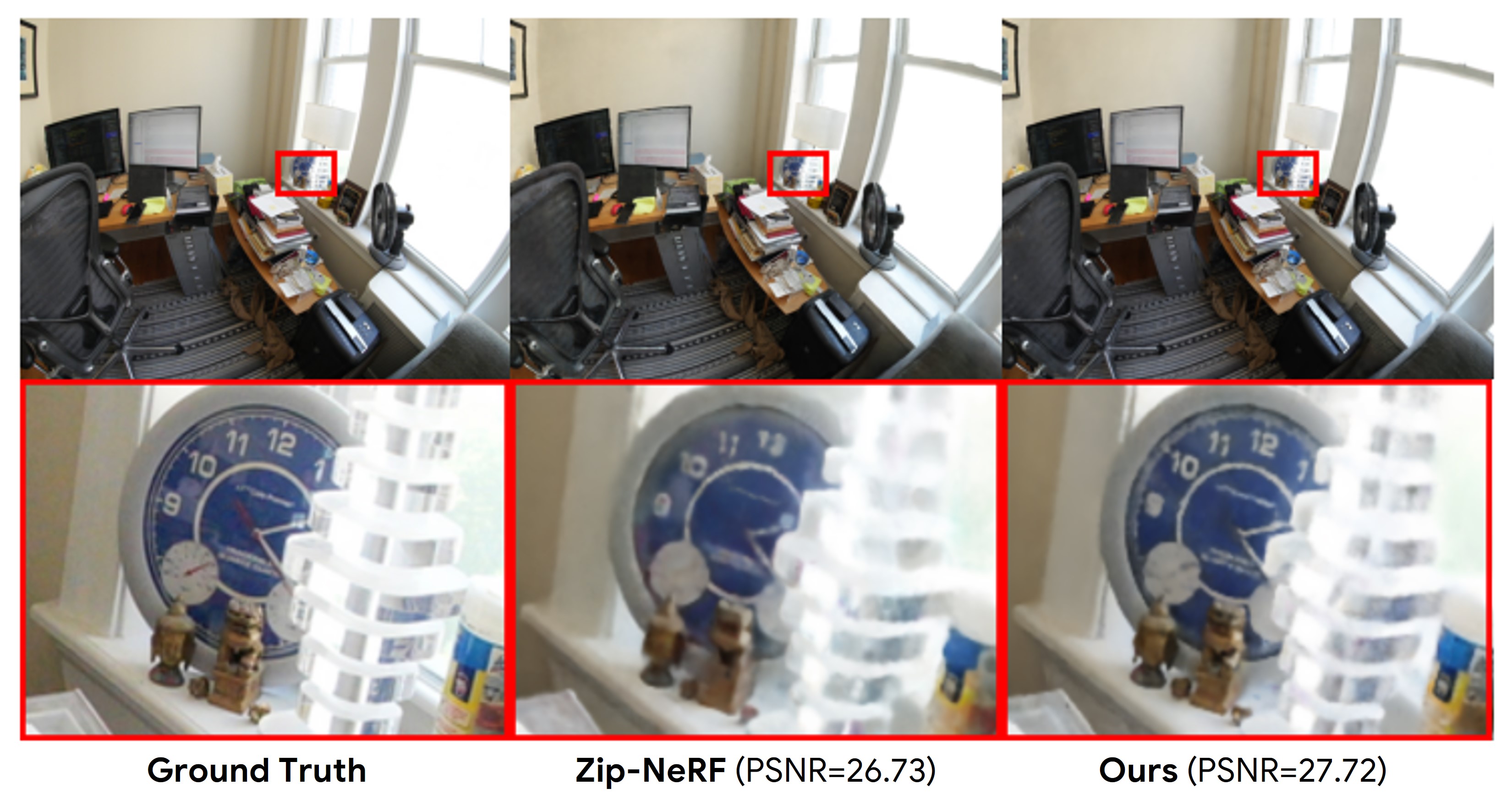}
  \end{subfigure}
  \vspace{-3pt}
  
  \caption{
      Images from \textsc{Berlin} and \textsc{NYC} in the \ZipnerfDataset, rendered by \ZipNeRF and \OurMethod with a 5$\times$4 grid.
  }
  \label{fig:comparison_zipnerf}
\end{figure}

%% file: figures/gardenvase_compare.tex
\begin{figure*}[ht]
  \centering
  \begin{subfigure}{\linewidth}
  \includegraphics[width=\textwidth]{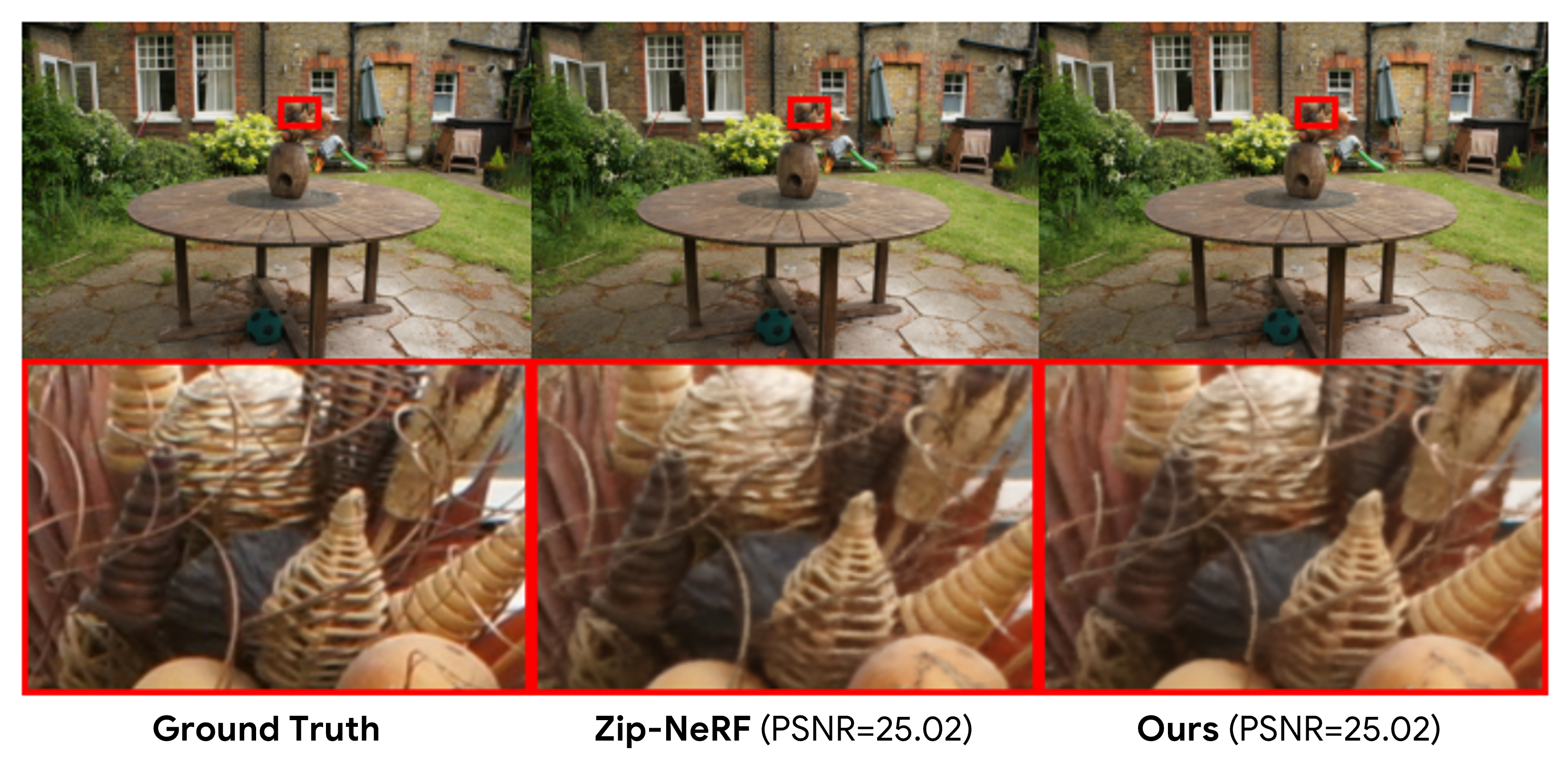}
  \end{subfigure}
  \hfill
  
  \caption{
      Renders of the \textsc{garden} scene in the \MipNeRF{} dataset using \ZipNeRF and \OurMethod with a 3$\times$3 parameter grid.
  }
  \label{fig:gardenvase-compare}
\end{figure*}

%% file: figures/qualitative.tex
\begin{figure*}[ht!]
  \centering
  \includegraphics[width=\textwidth]{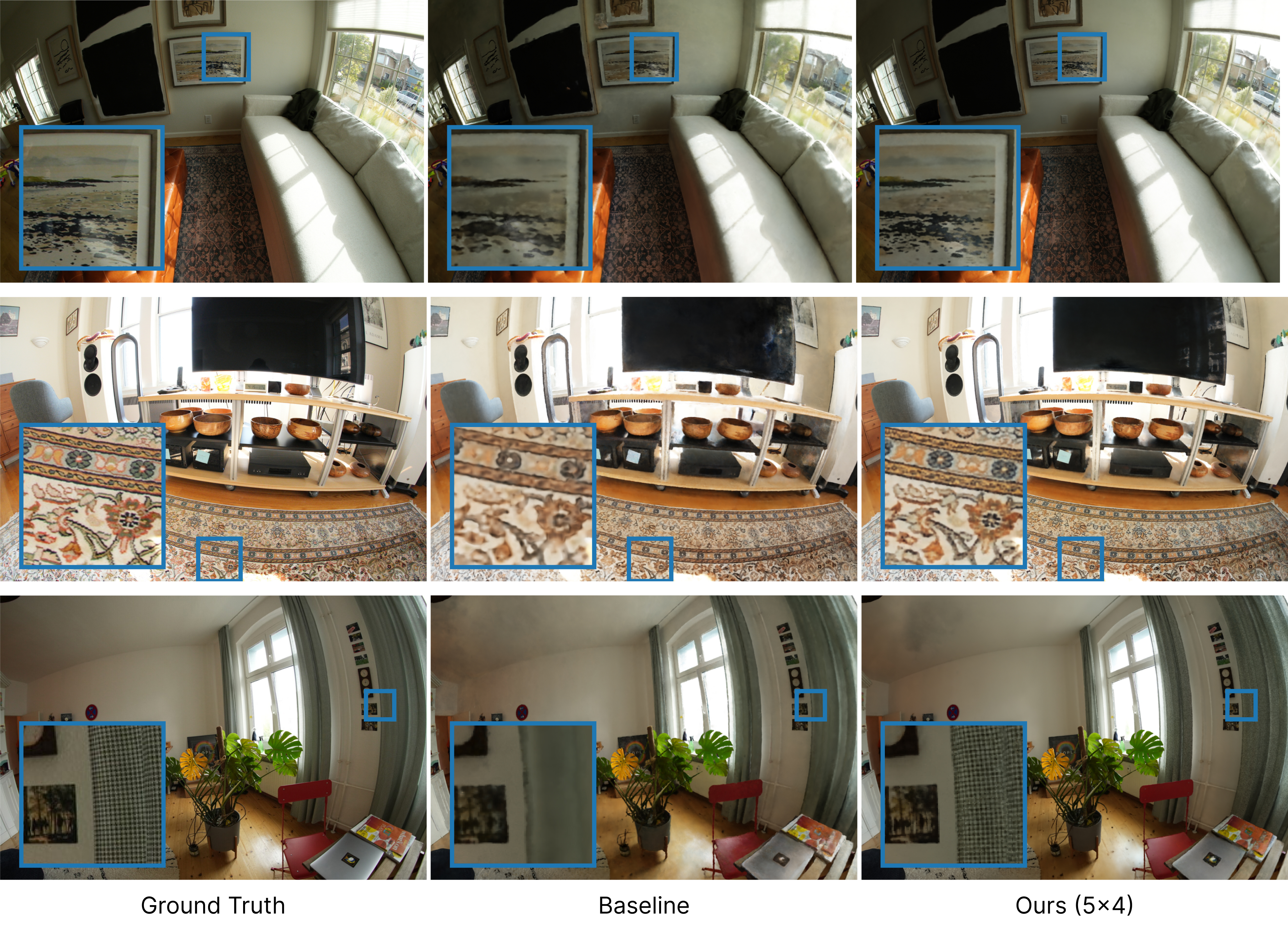}
  \caption{
    Additional qualitative results on the \ZipNeRF{} dataset.
    Our method consistently reconstructs centimeter-level texture detail in large, multi-room scenes.
  }
  \label{fig:qualitative-zipnerf}
\end{figure*}